\documentclass[letterpaper, 10 pt, conference]{ieeeconf}  

\IEEEoverridecommandlockouts                              

\title{\LARGE \bf
Combining Neural Networks and Tree Search for Task and Motion Planning in Challenging Environments
}

\author{Chris Paxton$^{1,2}$, Vasumathi Raman$^{1}$, Gregory D. Hager$^{2}$, Marin Kobilarov$^{1,2}$%
\thanks{$^{1}$Zoox, Inc., Menlo Park, CA, USA
        {\tt\small \{chris.paxton,vasu,marin\}@zoox.com}}
\thanks{$^{2}$The Johns Hopkins University, Baltimore, MD, USA
        {\tt\small cpaxton@jhu.edu, hager@cs.jhu.edu}}
}

\usepackage{times}
\usepackage{helvet}
\usepackage{courier}
\usepackage{latexsym}
\usepackage[english]{babel}
\usepackage{verbatim}
\usepackage{amsmath}
\usepackage{amsfonts}
\usepackage{relsize}
\usepackage{amssymb}  
\usepackage{algorithm}
\usepackage{algpseudocode}
\usepackage{epstopdf}
\usepackage{listings}
\usepackage{longtable}
\usepackage{textcomp}
\usepackage{url}
\usepackage{alltt}
\usepackage{caption}
\usepackage{subcaption}
\usepackage{graphicx}
\usepackage{color}
\usepackage{cite}

\DeclareMathOperator{\F}{\rotatebox[origin=c]{45}{$\Box$}}
\DeclareMathOperator{\G}{\Box}
\DeclareMathOperator{\X}{\bigcirc}
\DeclareMathOperator*{\argmax}{arg\,max}

\newtheorem{theorem}{Theorem}

\newtheorem{definition}[theorem]{Definition}

\makeatletter
\let\NAT@parse\undefined
\makeatother
\usepackage[numbers]{natbib}
\usepackage{multicol}
\usepackage[bookmarks=true]{hyperref}

\newcommand{\eat}[1]{}

\begin{document}

\maketitle
\thispagestyle{empty}
\pagestyle{empty}


\begin{abstract}

We consider task and motion planning in complex dynamic environments for problems expressed in terms of a set of Linear Temporal Logic (LTL) constraints, and a reward function. We propose a methodology based on reinforcement learning that employs deep neural networks to learn low-level control policies as well as task-level option policies. A major challenge in this setting, both for neural network approaches and classical planning, is the need to explore future worlds of a complex and interactive environment. To this end, we integrate Monte Carlo Tree Search with hierarchical neural net control policies trained on expressive LTL specifications. This paper investigates the ability of neural networks to learn both LTL constraints and control policies in order to generate task plans in complex environments. We demonstrate our approach in a simulated autonomous driving setting, where a vehicle must drive down a road in traffic, avoid collisions, and navigate an intersection, all while obeying given rules of the road.
\end{abstract}


\IEEEpeerreviewmaketitle

\section{Introduction}
A robot operating in the physical world must reason in a hybrid space: both its continuous motion in the physical world and the discrete goals it must accomplish are pertinent to correctly completing a complex task.
Common practice is to first compute a discrete action plan, then instantiate it depending on what is physically feasible. The field of Task and Motion Planning (TAMP) seeks to integrate the solving of the continuous and discrete problems since a robot's immediate physical motion is inextricably coupled with the discrete goal it seeks. One particular case where TAMP is particularly relevant is in the domain of autonomous driving. Self-driving cars have to deal with a highly complex and dynamic environment: they share a road with other moving vehicles, as well as with pedestrians and bicyclists. Road conditions are also unpredictable, meaning that such methods must be capable of dealing with uncertainty.

Current TAMP approaches combine high-level, ``STRIPS''-style logical planning with continuous space motion planning. These methods succeed at solving many sequential path planning and spatial reasoning problems~\cite{shivashankar2014towards,toussaint2015logic}, but dealing with dynamic environments and complex constraints is still a challenge~\cite{plaku2015motion}.
The problem is that the combined discrete and continuous state space tends to explode in size for complex problems. The addition of temporal constraints makes the search problem even more difficult, though there has been recent progress in this direction~\cite{plaku2015motion}.

\begin{figure}[bt]
\centering
\includegraphics[width=0.99\columnwidth]{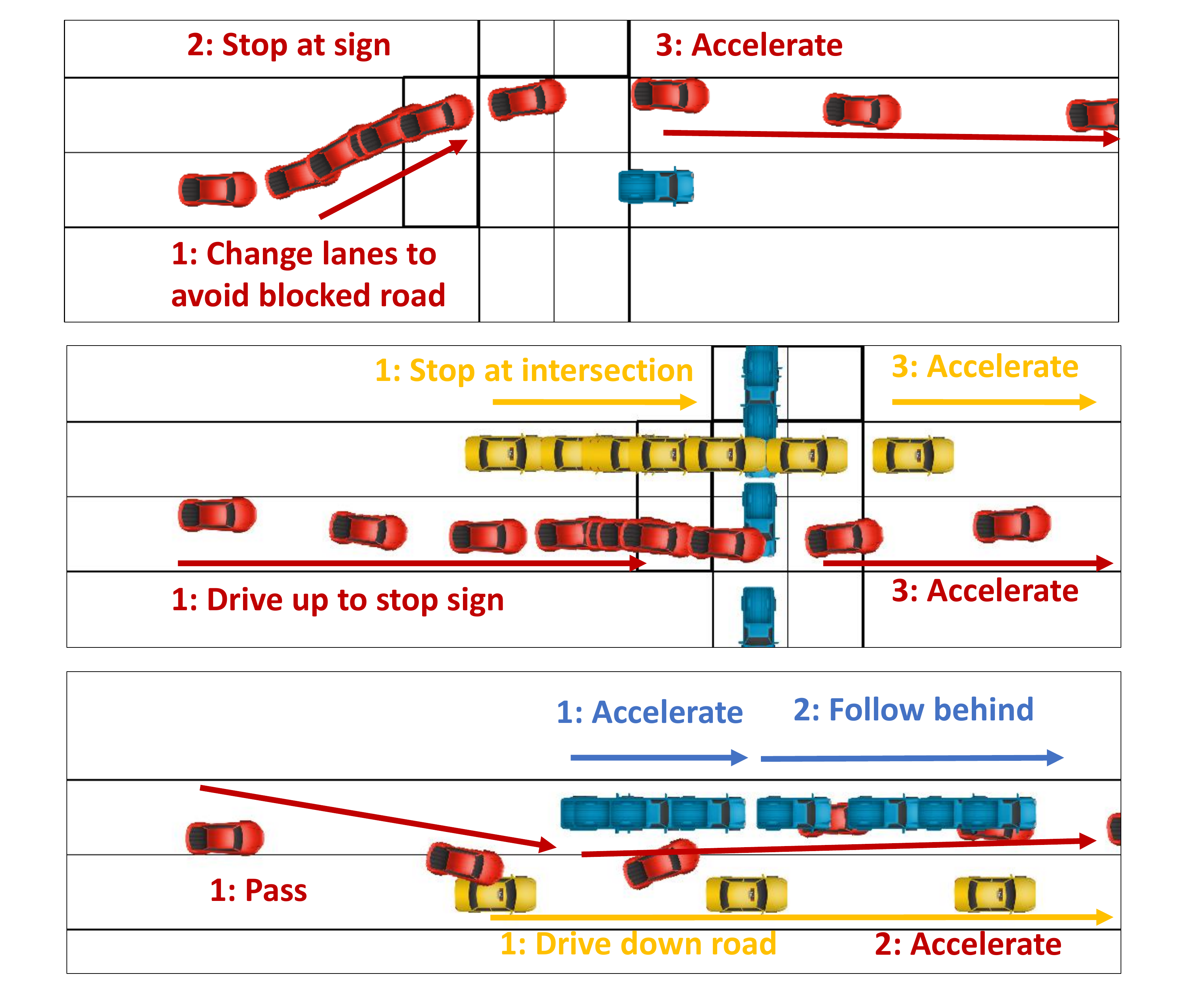}
\caption{Simulated self driving car problems containing an intersection and multiple vehicles. The car must be able to behave intelligently when confronted with sudden obstacles like stopped or slow moving vehicles, and it must be able to obey the rules of the road.}
\label{fig:driving-task}
\end{figure}

On the other hand, recent work in Deep Reinforcement Learning (DRL) has shown promise in control policy learning in challenging domains~\cite{lillicrap2015continuous,silver2016mastering,mnih2016asynchronous,bojarski2016end}, including robotic manipulation~\cite{el2017deep} and autonomous driving~\cite{bojarski2016end,xu2016end,shalev2016safe}.
DRL has also been combined with Monte Carlo Tree Search (MCTS) for learning~\cite{guo2014deep} and game playing~\cite{silver2016mastering} in a variety of discrete and continuous environments. 
However, the question remains open whether these approaches can be integrated in a TAMP framework to produce reliable robot behavior.

In this paper, we show that the best of both worlds can be achieved by using neural networks to learn both low-level control policies and high-level action selection priors, and then using these multi-level policies as part of a heuristic search algorithm to achieve a complex task. 
A major challenge of complex, dynamic environments is that there is no straightforward heuristic to guide the search towards an optimal goal: since the goals are temporally specified, there is no way to anticipate constraints and conflicts that may arise further on in the plan.
To address this issue, we formulate task and motion planning as a variant of Monte Carlo Tree Search over high-level options, each of which is represented by a learned control policy, trained on a set of LTL formulae. The key to performance with MCTS is to start with a good option selection policy, which in this work is provided by a learned high-level policy over discrete options.
This approach allows us to efficiently explore the relevant parts of the search space to find high quality solutions when other methods would fail to do so.

To summarize, the contributions of this paper are:
\begin{itemize}
\item A planning algorithm combining learned low-level control policies with learned high level ``option policies'' over these low-level policies for TAMP in dynamic environments.
\item A framework for incorporating complex task requirements expressed in temporal logic
\item Evaluation of our approach in a simulated autonomous driving domain.
\end{itemize}

Note that while our approach performs very well in simulation, it still has several limitations, which we discuss in Section \ref{sec:conc}.

\section{Preliminaries}\label{sec:prelims}
In this section we establish notation and terminology for the system modeling and task specification formalisms, Markov Decision Processes (MDPs) and linear temporal logic (LTL), respectively.

\subsection{System model}
In keeping with most reinforcement learning algorithms, we model the system under consideration as a Markov Decision Process (MDP) \cite{bellman1967}. Learning is performed over a sequence of time steps. At each step $t$, the agent observes a state, $s_t \in S$, which represents the sensed state of the system, i.e., its internal state as well as what it perceives about the environment it operates in. 
\eat{Note that $S$ can be defined to include dynamic and kinematic models of the agent and environment.} Based on $s_t$, the agent selects an action $a_t \in A$ from an available set of actions. On performing $a_t$ in $s_t$, the agent receives an immediate reward, $r_t \in R$, and moves to a state in set $s_{t+1} \in \delta(s_t,a_t)$. The goal of the agent is to maximize its cumulative reward or a time-discounted sum of rewards over a time horizon (which may be finite or infinite). Without loss of generality, the agent acts according to a policy, $\pi: S \rightarrow A$. A \emph{run} of an MDP ${\bf s} = s_0s_1s_2 \cdots$ is an infinite sequence of states such that for all $t$, exists $a_t \in A$ such that $s_{t+1} \in \delta(s_t,a_t)$.

\eat{A useful feature of an MDP, both for dynamic programming and learning-based methods, is the \emph{$Q$-function}. }Given a current state $s$, the value of $Q^\pi(s,a)$ is defined to be the best cumulative reward that can be obtained in the future under policy $\pi$ after performing action $a$.  The $Q$-function is thus a local measure of the quality of action $a$. Similarly, the \emph{value function} of an MDP $V^\pi:S \rightarrow R$ is a local measure of the quality of $s$ under policy $\pi$. For an optimal policy $\pi^*$, $V^*$ and $Q^*$ are obtained as fixed points using Bellman's equation. Most reinforcement learning algorithms approximate either the the $V$ function or the $Q$ function. 
\eat{In particular, actor-critic policy iteration methods aim to learn a policy in an iterative manner, where for each iteration $i$ the ``critic'' estimates $Q^{\pi_i}$, and the ``actor'' uses this to improve $\pi_i$ to get $\pi_{i+1}$.} For more detail, the interested reader is referred to the survey of RL methods by \cite{koberBP13}.

To solve the MDP, we pick from a hypothesis class of policies $\pi$ composed using a set of high-level options, which are themselves learned from a hypothesis class of parametrized control policies  using a deep neural network. As such, the optimal policy may not be contained in this hypothesis class, but we are able to demonstrate architectures under which a good approximation is obtained. 

One key challenge for RL methods is the requirement of an agent model that is Markovian. In this work, we augment the state space $S$ with memory, in the form of a deterministic Rabin automaton obtained from an LTL specification. This is similar to approaches in the formal methods literature like \cite{DingSBR14} which construct a product of an MDP and a Rabin automaton. However, in contrast with these approaches, we do not obtain an optimal MDP policy via dynamic programming, but approximate it via deep reinforcement learning.

\subsection{Linear Temporal Logic}

We prescribe properties of plans in terms of a set of atomic statements, or propositions. An atomic proposition is a statement about the world that is either \texttt{True} or \texttt{False}. Let $AP$ be a finite set of atomic propositions, indicating properties such as occupancy of a spatial region, and a labeling function ${\cal L} : S \rightarrow 2^{AP}$ a map from system states to subsets of atomic propositions that are \texttt{True} (the rest being false). For a given run $\bf s$, a \emph{word} is the corresponding sequence of labels ${\cal L}({\bf s}) = {\cal L}(s_0) {\cal L}(s_1) {\cal L}(s_2) \cdots$.
\eat{A \emph{word} is an infinite sequence of labels ${\cal L}({\bf s}) = {\cal L}(s_0) {\cal L}(s_1) {\cal L}(s_2) \cdots$, for some run ${\bf s}$. Let ${\bf s}_i = s_is_{i+1}s_{i+2}\cdots$ denote the suffix of ${\bf s}$ starting at index $i$, with corresponding word ${\cal L}({\bf s}_i)$} 
We use linear temporal logic (LTL) to concisely and precisely specify permitted and prohibited system behaviors in terms of the corresponding words. \eat{LTL is an expressive language that has been used to specify a wide range of system behaviors for robots.} We briefly review the syntax and semantics of LTL, and refer the interested reader to \cite{MCBk} for further details.

\noindent \textbf{Syntax:} Let $AP$ be a set of atomic propositions. Formulas are constructed from $p \in AP$ according to the grammar:
$$\varphi ::=~p~|~\neg \varphi~|~\varphi_1 \lor \varphi_1~|~\X \varphi~|~\varphi_1~{\mathcal U}~\varphi_2$$
where $\neg$ is negation, $\vee$ is disjunction, $\X$ is ``next" , and ${\mathcal U}$ is ``until". 
Boolean constants $\mathtt{True}$ and $\mathtt{False}$ are defined as usual: $\mathtt{True} = p \vee \neg p$ and $\mathtt{False} = \neg\mathtt{True}$. Conjunction ($\wedge$), implication ($\Rightarrow$), equivalence ($\Leftrightarrow$), ``eventually" ($\F \varphi = \mathtt{True}~{\mathcal U}~\varphi$) and ``always" ($\G \varphi = \neg \F \neg \varphi$) are derived.\\
\noindent \textbf{Semantics:}
\eat{The semantics of LTL are defined inductively
over a word ${\cal L}({\bf s})$ as follows:
\[
\begin{array}{lll}
{\cal L}({\bf s}_i) &\models& p \text{ if and only if } p \in {\cal L}({\bf s}_i) \\
{\cal L}({\bf s}_i) &\models& \neg \varphi \text{ if and only if } {\cal L}({\bf s}_i) \not \models \varphi \\
{\cal L}({\bf s}_i) &\models& \varphi_1 \lor \varphi_2 \text{ if and only if } {\cal L}({\bf s}_i) \models \varphi_1 \\
&&~~~~\text{ and } {\cal L}({\bf s}_i) \models \varphi_2 \\
{\cal L}({\bf s}_i) &\models& \X \varphi \text{ if and only if } {\cal L}({\bf s}_{i+1}) \models \varphi \\
{\cal L}({\bf s}_i) &\models& \varphi_1~{\cal U}~\varphi_2 \text{ if and only if } \exists j \ge i \text{ such that } \\
&&~~~~{\cal L}({\bf s}_j) \models \varphi_2 \text{ and } \forall i \le k \le j,~{\cal L}({\bf s}_k) \models \varphi_1
\end{array}
\]
A word ${\cal L}({\bf s})$ satisfies $\varphi$, denoted by ${\cal L}({\bf s}) \models \varphi$, if ${\cal L}({\bf sx}_0) \models \varphi$. A run $\bf s$ satisfies $\varphi$ if ${\cal L}({\bf s}) \models \varphi$. 
}
%
Informally, $\X \varphi$ expresses that $\varphi$ is true in the next ``step" or position in the word, $\varphi_1 ~{\mathcal U}~ \varphi_2$ expresses that $\varphi_1$ is true until $\varphi_2$ becomes true, $\G \varphi$ means that $\varphi$ is true in every position, $\F\varphi$ means $\varphi$ is true at some position, and $\G\F\varphi$ means $\varphi$ is true infinitely often (it reoccurs indefinitely). A run $\bf s$ satisfies $\varphi$ (denoted by ${\bf s} \models \varphi$) if and only if the word ${\cal L}({\bf s})$ does.

\eat{LTL provides the ability to describe changes in the atomic propositions over time.}
To allow users who may be unfamiliar with LTL to define specifications, some approaches such as that of \cite{RamanLFLMG13} include a parser that automatically translates English sentences into LTL formulas. Many applications distinguish two primary types of properties allowed in a specification -- \emph{safety} properties, which guarantee that ``something bad never happens", and \emph{liveness} conditions, which state that ``something good (eventually) happens". These correspond naturally to LTL formulas with operators ``always'' ($\G$) and ``eventually'' ($\F$). 

Another useful property of LTL is the equivalence between LTL formulas and Deterministic Rabin Automata (DRAs).
\eat{
\begin{definition}[Deterministic Rabin automata]
A \emph{deterministic Rabin automaton} is a tuple ${\cal A }= (Q, \Sigma, \delta, q_0, {\cal F})$ consisting of 
\begin{enumerate}
\item a finite set of states $Q$
\item a finite alphabet $\Sigma$
\item a transition function $\delta: Q \rightarrow Q$
\item an initial state $q_0 \in Q$, and
ou\item a set of accepting pairs $\Omega = \{(L_1, U_1), \cdots, (L_N, U_N)\}$.
\end{enumerate}
\end{definition}
Let $\Sigma^\omega$ be the set of infinite words over $\Sigma$. A \emph{run} of $\cal A$ is an infinite sequence $q_0q_1q_2\cdots$ of states in $\cal A$ such that there exists a \emph{word} ${\sigma} = \sigma_0\sigma_1\sigma_2\cdots \in \Sigma^\omega$ with $(q_i,\sigma_i) = q_{i+1}$ for $i \ge 0$. Run $q_0q_1q_2 \cdots$ is \emph{accepted} by ${\cal A}$ if there is a pair $(L_j , U_j ) \in \Omega$ such that $q_i \in L_j$ for infinitely many indices $i \in \mathbb{N}$ and $q_i \in U_j$ for at most finitely many $i$.

Denote by ${\cal L}({\cal A})$ the set of words that are accepted by $\cal A$. Any LTL formula $\varphi$ over variables in $AP$ can be automatically translated into a corresponding DRA ${\cal A}_\varphi$ of size $2^{2^{|AP|}}$ such that $\sigma \in {\cal L}({\cal A}_\varphi) \iff \sigma \models \varphi$. 
}
Any LTL formula $\varphi$ over variables in $AP$ can be automatically translated into a corresponding DRA ${\cal A}_\varphi$ of size automaton $2^{2^{|AP|}}$ that accepts all and only those words that satisfy $\varphi$ \cite{MCBk}.

\section{Approach}\label{sec:approach}
We consider systems that evolve according to continuous dynamics $f_c$ and discrete dynamics $f_d$:
\[
x' = f_c(x,w,u,o),~w' = f_d(x,w,u,o)
\]
where
\begin{itemize}
\item $x \in {\cal X} \subseteq \mathbb{R}^{n}$ is the continuous state
\item $u \in {\cal U} \subseteq \mathbb{R}^{m}$ is the continuous control input
\item $w \in {\cal W}$ is the discrete (logical) world state 
\item $o \in {\cal O}$ is a discrete (logical) option from a finite set $\cal O$
\end{itemize}

Our atomic propositions $p \in AP$ are defined as functions over the discrete world state, i.e. $p: {\cal W} \rightarrow \{\texttt{True},\texttt{False}\}$. 

In the MDP framework, $S = {\cal X} \times {\cal W}, A = {\cal U} \times {\cal O}, \delta(xw, uo) = x'w' \text{ such that } x' = f_c(x,w,u,o), ~ w' = f_d(x,w,u,o)$. The labeling function over states is ${\cal L}(xw) = \{p \in AP \text{ such that } p(w)=\texttt{True}\}$.

We decompose the system into many actors. Each independent entity is an actor, and in particular the agent under control is an actor. A world state $s = xw \in {\cal X} \times {\cal W}$ consists of an environment $e \in \mathcal{E}$ and some number of actors $N$. The $i$-th world state in a sequence is therefore fully defined as:
\[
	x_iw_i = \left< x_{0,i}w_{0,i}, x_{1,i}w_{1,i}, \dots, x_{N,i}w_{N,i}, e \right>,
\]
where each actor $k$'s state $x_{k,i} \in \mathbb{R}^{n_{}}$ and $w_{k,i} \in {\cal W}_k$ such that $\sum_k n_{k} = n$ and $\prod {\cal W}_k = {\cal W}$, and actor $0$ is the planner.
Actors represent other entities in the world that will update over time according to some unspecified policy specific to them.

Finally, we assume we are given a feature function $\phi: S \rightarrow {\cal F}$, which computes a low-dimensional representation of the world state containing all  information needed to compute a policy. 
For example, when doing end-to-end visuomotor learning as in~\cite{bojarski2016end,xu2016end}, $\phi(s)$ would simply return the camera image associated with that particular world state. We show other examples of this feature function in our experiments.

We seek to learn a set of behaviors that allow an actor to plan safe, near-optimal trajectories through the environment out to a fixed time horizon while obeying a set of discrete constraints.
We decompose this problem into finding two sets of policies: a policy $\pi_\mathcal{O}:{\cal F} \rightarrow {\cal O}$ over high-level actions and a policy $\pi_\mathcal{U}:{\cal O} \times {\cal F} \rightarrow {\cal U}$ over low-level controls, such that their composition solves the MDP. In particular, our first subgoal is to compute a policy $\pi_{\mathcal{U}}^*(\cdot, o)$ for each high-level option $o$ that maps from arbitrary feature values to controls:
\[
\pi_{\mathcal{U}}^*(\phi(xw),o) = \argmax_u (V^*(\delta(xw,uo)))
\]

We also compute a second policy over options, $\pi_{\mathcal{O}}^*$:
\[
\pi_{\mathcal{O}}^*(\phi(xw)) = \argmax_o (V^*(\delta(xw,\pi_{\mathcal{U}}^*(\phi(xw),o)o)))
\]

\begin{figure*}[bt]
\includegraphics[width=2\columnwidth]{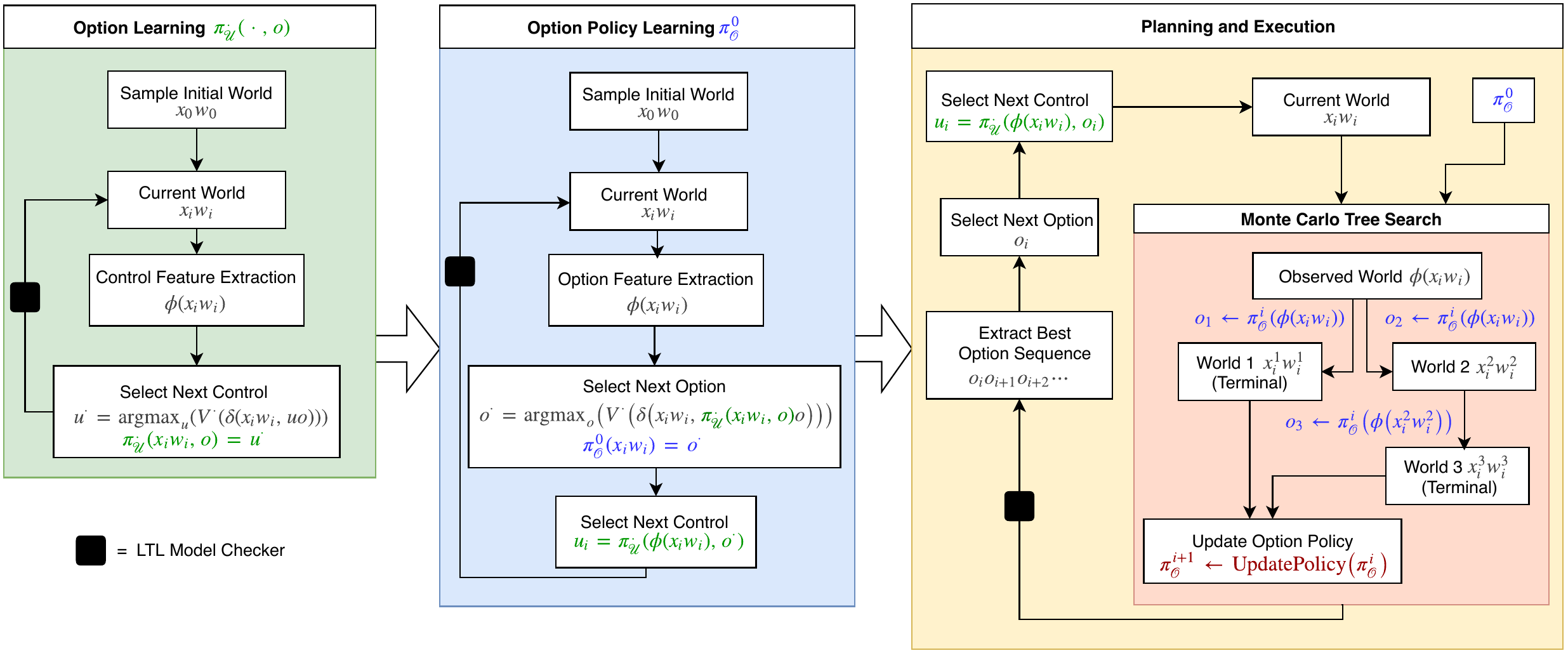}
\caption{Workflow diagram of the proposed approach, showing the training process and planning/execution loop.}
\label{fig:system}
\end{figure*}

This high-level policy tells us what we expect to be the best control policy to execute over some short time horizon.
Note that since we are imposing additional structure on the final policy (which takes the form $\pi^*(s) = \pi_{\mathcal{U}}^*(\phi(s),\pi_{\mathcal{O}}^*(\phi(s)))$), it may no longer be truly optimal. However, it is the optimal such policy found with this set of options $\cal O$. Our results show that decomposing the problem in this way, by learning a simple set of options to plan with, is effective in the autonomous driving domain. Without leveraging the inherent structure of the domain in this manner, end-to-end training would learn $\pi^*(s)$ directly, but require a much more sophisticated training method and a lot more data.
Fig.~\ref{fig:system} provides a schematic overview of our proposed approach.

\subsection{Planning Algorithm}\label{sec:algorithm}

In a dynamic environment with many actors and temporal constraints, decomposing the problem into reasoning over goals and trajectories separately as in prior work~\cite{toussaint2015logic,shivashankar2014towards} is infeasible.
Instead, we use learned policies together with an approach based on MCTS.
The most commonly used version of MCTS  is the Upper Confidence Bound (UCB) for Trees.
We recursively descend through the tree, starting with $s = s_0$ as the current state. At each branch we would choose a high level option according to the UCB metric:
\[
	Q(s_i,o_i) = Q^*(s_i,o_i) + C \sqrt{\dfrac{\log(N(s_i, o_i))}{N(s_i) + 1}}
\]
where $Q^*(s_i,o_i)$ is the average value of option $o_i$ from simulated play, $N(s_i, o_i)$ is the number of times option $o_i$ was observed from $s_i$, and $N(s_i)$ is the number of times $s_i$ has been visited during the tree search.
$C$ is an experimentally-determined, domain-specific constant.

Our particular variant of MCTS has two specializations. First, we replace the usual Upper Confidence Bound (UCB) weight with the term from~\cite{silver2016mastering} as follows:
\[
	Q(s_i,o_i) = Q^*(s_i,o_i) + C \dfrac{P(s_i, o_i)}{1 + N(s_i, o_i)}
\]
where $p(s_i, o_i)$ is the predicted value of option $o_i$ from state $s_i$.
The goal of this term is to encourage exploration while focusing on option choices that performed well according to previous experience; it grants a high weight to any terms that have a high prior probability from our learned model.

Next, we use Progressive Widening to determine when to add a new node.
This is a common approach for dealing with Monte Carlo tree search in a large search space~\cite{couetoux2011continuous,yee2016monte}.
It limits the maximum number of children of a given node to some sub-linear function of the number of times a world state has been visited, commonly implemented as
\[
	N_{children}(s_i)^* = (N(s_i))^\alpha
\]
with $\alpha \in (\frac{1}{2}, \frac{1}{4})$.
We use Progressive Widening together with our learned policy to reduce the number of trees that must be visited to generate a plan.
Whenever we add a new node to the search tree, we additionally use the current high-level policy to explore until we reach a termination condition.

After selecting an option to explore, we call the \textsc{simulate} function to evolve the world forward in time. During this update, we check the full set of LTL constraints $\Phi$ and associated option constraints $\varphi_o$. If these are not satisfied, the search has arrived in a terminal node and a penalty is applied.
Alg.~\ref{alg:complete} is the standard algorithm for Monte Carlo Tree Search with these modifications.

\begin{algorithm}[bt!]
\caption{Pseudocode for MCTS over options.}
\label{alg:complete}
\begin{algorithmic}
\small
\Function{select}{$s$}
\If{$s$ is terminal}
  \Return $v(s)$
\Else
  \If{$N(s) = 0$ or $N_{children}(s) < \sqrt{N(s)}$}
    \State Add new edge with option $o$ to children
  \EndIf    
  \State $o^* = \argmax_o$
  \If {$N(s, o^*) = 0$}
  	\State // This branch has not yet been explored
  	\State $s' =$ \Call{Simulate}{$s$,$o^*$}
  \Else
  	\State // This branch has been previously simulated
  	\State $s' =$ \Call{Lookup}{$s$,$o^*$}
  \EndIf
  \State // Recursively explore the tree
  \State $v' =$ \Call{Select}{$s'$}
  \State // Update value associated with this node
  \State $v = v(s) + v'$
  \State \Call{Update}{$v$}
  \State \Return $v$
\EndIf
\EndFunction
\Function{simulate}{s,o}
\While{$t < t_{max}$ and $c(o)$}
\State $u^* = \pi_{\mathcal{U}}^*(\phi(s), o)$
\State $s' = $\Call{advance\_world}{$s$, $u^*$, $\Delta t$}
\State $t = t + \Delta t$
\For {$\varphi \in \Phi \cup \varphi_o$}
\If {${\bf s} \not\models \varphi$}
	\State $s' \leftarrow terminal$
	\State \Return $s'$
\EndIf
\EndFor
\EndWhile
\State \Return $s'$
\EndFunction
\end{algorithmic}
\end{algorithm}

\subsection{Model Checking}\label{sec:model_checking}
Each discrete option is associated with an LTL formula $\varphi_o$ which establishes conditions that must hold while applying that option. We can evaluate $u_i = \pi_{\cal U}(o,\phi(x_iw_i))$ to get the next control as long as $\varphi_o$ holds. In addition, we have a shared set $\Phi$ of LTL formulae that constrain the entire planning problem. 

In order to evaluate the cost function when learning options, as well as during MCTS, we check whether sampled runs satisfy an LTL formula. Since we are checking satisfaction over finite runs, we use a bounded-time semantics for LTL \cite{BiereHJLS06}. We precompute maximal accepting and rejecting strongly connected components of the DRA, which enables model checking in time linear in the length of the run.

\section{Self Driving Car Domain}

We apply our approach to the problem of planning for a self-driving car passing through an all-way stop intersection. 
To successfully complete the task the car must:
(1) accelerate to a reference speed,
(2) stop at a stop sign,
(3) wait until its turn to move,
(4) accelerate back to the reference speed,
all while avoiding collisions and changing lanes as necessary.
We break this down into a set of mid-level options:
\begin{align*}
\mathcal{O} = \{&\mathtt{Default}, \mathtt{Follow}, \mathtt{Pass}, \mathtt{Stop},
\\&\mathtt{Wait}, \mathtt{Left}, \mathtt{Right}, \mathtt{Finish}\}
\end{align*}
where the \texttt{Default} option represents driving down a straight road and stopping at a stop sign. 
The \texttt{Wait} option defines behavior at an intersection with multiple other vehicles, and captures behavior where the agent waits until its turn before moving through the intersection. Other options such as \texttt{Stop} and \texttt{Finish} represent the agent's behavior on a crowded road before and after the intersection, respectively.

The \texttt{Default} option was trained on roads with random configurations of vehicles, but with the agent's own lane clear. This option learns lane keeping behavior and LTL constraints. We also learn a \texttt{Stop} agent, which terminates when stopped at a stop region, and a \texttt{Wait} agent, which starts in a stop region where multiple other vehicles have priority and must pass through the intersection successfully. The \texttt{Left} and \texttt{Right} options are successful if they move entirely into the other lane within 21 meters. We also train \texttt{Follow} and \texttt{Pass} options on scenarios where another car is immediately ahead of the agent. In \texttt{Follow}, the agent is restricted to its initial lane; in \texttt{Pass} it is not.

\subsection{Vehicle Modeling}

We employ a non-slip second-order nonholonomic model that is often sufficient for realistic motion planning in nominal driving conditions. 
%
The physical state of each vehicle entity in the world is defined by $x = (p_x, p_y, \theta, v, \psi)$, where $(p_x, p_y)$ denotes the inertial position at the rear axle, $\theta$ is its heading relative to the road, $v$ is the velocity and $\psi$ is the steering angle. The control inputs are the acceleration $a$ and steering angle rate $\dot \psi$, i.e. $u = (u_1,u_2) := (a, \dot\psi)$. The dynamics of all vehicles are defined as
\begin{align*}
    &    \dot p_x = v \cos\theta, \\
	&\dot p_y = v \sin\theta, \\
    &\dot \theta = v \frac{\tan\psi}{L},\\
    &\dot v = u_1, \\
    &\dot \psi = u_2,
\end{align*}
where $L$ is the vehicle wheel-base. These equations are integrated forward in time using a discrete-time integration using a fixed time-step $\Delta t = 0.1$ seconds during learning and for all experiments.

\subsection{Road Environment}

Our scenarios take place at the intersection of two two-lane, one-way roads. We choose this setting specifically because it creates a number of novel situations that the self-driving vehicle may need to deal with. Each lane is 3 meters wide with a 25 mph speed limit, corresponding to common urban or suburban driving conditions.
The target area is a 90 meter long section of road containing an intersection, where two multi-lane one-way roads pass through each other. Stop signs are described as ``stop regions'': areas on the road that vehicles must come to a stop in before proceeding.

Other vehicles follow an aggressive driving policy. If far enough away from an event, they will accelerate up to the speed limit (25 mph). Otherwise, they will respond accordingly.
If the next event on the road is a vehicle, they will slow down to maintain a follow distance of 6 meters bumper to bumper.
They will decelerate smoothly to come to a stop halfway through a stop region. They will remain stopped until the intersection is clear and they have priority over other waiting vehicles. Once in an intersection, they will accelerate up to the speed limit until they are clear. They have a preferred acceleration of $1.0 m/s^2$, and will not brake harder than $2.0 m/s^2$. Actors will appear on the road moving at either the speed limit or a point on the reference velocity curve (if before a stop sign). 

\subsection{Cost and Constraints}\label{sec:LTL}
As described in Section \ref{sec:model_checking}, each discrete option is associated with an LTL formula $\varphi_o$ which establishes conditions that must hold while applying it, and we also have a shared set $\Phi$ of LTL formulae constraining the entire plan.

For example, the \texttt{Wait} option learns to wait at an intersection, and then pass through that intersection, so $\varphi_{\texttt{Wait}}=$
\[
\begin{array}{l}
\G (\text{has\_stopped\_in\_stop\_region} \Rightarrow \\ ~~~\left(\text{in\_stop\_region} \vee \text{in\_intersection} \right))
\end{array}
\]

For the road scenario, $\Phi =$
\[
\begin{array}{l}
\G \left(\text{in\_stop\_region} \Rightarrow \left(\text{in\_stop\_region}\right.\right. \\
~~~~~~~~~~~~~~~~~~~~~~~~~{\cal U}~\left.\left.\text{has\_stopped\_in\_stop\_region}\right)\right)\\ 
\G ((\text{in\_intersection} \Rightarrow \text{intersection\_is\_clear}) \land\\
~~~~~~~~~~~~~\neg \text{in\_intersection}~{\cal U}~\text{higher\_priority})
\end{array}
\]

The reward function is a combination of a a cost term based on the current continuous state and a bonus based on completing intermediate goals or violating constraints (e.g. being rejected by the DRA corresponding to an LTL formula).
The cost term penalizes the control inputs, acceleration and steering angle rate, as well as jerk, steering angle acceleration, and lateral acceleration. There are additional penalties for being far from the current reference speed and for offset from the center of the road. We add an additional penalty for being over the reference speed, set to discourage dangerous driving while allowing some exploration below the speed when interacting with other vehicles.
This cost term is:
\[
   r_{cost} = \|(e_y, e_\theta, v - v_{ref}, \min(0, v_{ref} - v), a, \dot a, \dot\psi) \|^2_W,
\]
expressed as the weighted $L_2$ norm of the residual vector with respect to a diagonal weight matrix $W$. Here the errors $e_y$ and $e_\theta$ encode the lateral error and heading error from the lane centerline.  

We add terminal penalties for hitting obstacles or violating constraints.
The other portion of the reward function are constraint violations. As noted by~\cite{shalev2016safe}, the penalty for rare negative events (and, correspondingly, the reward for rare positive events) must be large compared to other costs accrued during a rollout. We set the penalty for constraint violations to $-200$ and provide a $200$ reward for achieving goals: stopping at the stop sign and exiting the region. 
When training specific options, we set an additional terminal goal condition: these include stopping at a stop region for the \texttt{Stop} option, passing through an intersection for the \texttt{Wait} option, and changing lanes within a certain distance for the \texttt{Left} and \texttt{Right} options.
We found that this architecture was better able to learn the LTL constraints associated with the problem. Note that these terminal goal conditions correspond well to additional, option-specific LTL specifications (e.g. $\F\text{in\_stop\_region}$).

\subsection{Learning}\label{sec:learning}

Our approach for multi-level policy learning is is similar to the framework used by~\cite{andreas2016modular}, who use a multi-level policy that switches between multiple high-level options.
This allows us to use state of the art methods such as~\cite{lillicrap2015continuous,gu2016continuous} for learning of individual continuous option-level policies.
All control policies are represented as multilayer perceptrons with a single hidden layer of 32 fully connected neurons. We used the ReLu activation function on both the input and hidden layer, and the tanh activation function on the outputs. Outputs mapped to steering angle rate $\dot\psi \in [-1,1]$ rad/s and acceleration $a \in [-2,2]$ m/s$^2$.

These models were trained using Keras~\cite{chollet2015keras} and TensorFlow. We used the Keras-RL implementations of deep reinforcement learning algorithms DQN, DDPG, and continuous DQN~\cite{plappert2016kerasrl}. These were trained for 1 million iterations with the Adam optimizer, a learning rate of $1e-3$. Exploration noise was generated via an Ornstein-Uhlbleck process with sigma that annealed between 0.3 and 0.1 over 500,000 iterations.
We examined two different reinforcement learning approaches for finding the values of our low-level controllers: Deep Direct Policy Gradients (DDPG), as per~\cite{lillicrap2015continuous}, and Continuous Deep Q Learning (cDQN)~\cite{gu2016continuous}. However, cDQN had issues with convergence for some of our options, and therefore we only report results using low-level policies learned via DDPG.
It should be noted that this is a very difficult and complex optimization problem: the LTL constraints  and other actors in the same lane introduce sharp nonlinearities that do not exist when simply learning a controller to proceed down an otherwise empty road.

We then performed Deep Q learning~\cite{mnih2013playing} on the discrete set of options to learn our high-level options policy.
High-level policies were trained on a challenging road environment with 0 to 6 randomly placed cars with random velocity, plus a 50\% chance of a stopped vehicle ahead in the current lane.
We use Deep Q Learning as per~\cite{mnih2013playing} to learn a stochastic high-level policy over either these learned options, the manually defined ones, or a combination of the two.

We include features that capture the planning actor's heading and velocity on the road, as well as its relationship to certain other actors. We designed our set of features to be generalizable to situations with many or few other actors on the road. 
There is a fixed maximum horizon distance for distances to other entities in the world, with the maximum response at 0 distance and no response beyond this horizon.
The feature set for the current actor based on its continuous state is:
\[
	\phi_{x}(s) = \left[v, v_{ref}, e_y, \psi, v\frac{\tan\psi}{L}, \theta, lane, e_{y,lane}, a, \dot{\psi}\right],
\]
where we also included the previous control input $u=(a,\dot\psi)$ (for clarity of presentation $u$ is not explicitly included in the current state $s$). To this, we append the set of predicates:
\begin{align*}
	\phi_{w}(s) = [&\mathtt{not\_in\_stop\_region}, \\
	&\mathtt{has\_entered\_stop\_region},\\
	&\mathtt{has\_stopped\_in\_stop\_region},\\
	&\mathtt{in\_intersection},\mathtt{over\_speed\_limit},\\
	&\mathtt{on\_route},\mathtt{intersection\_is\_clear},\\
	&\mathtt{higher\_priority}]
\end{align*}

For the vehicles ahead, behind, in the other lane ahead and behind, and on the cross road to the left and right, we add the set of features related to that actor (assuming its index is $j$, while the index of the vehicle for which the policy is applied to be $i$): 
\[
	\phi_{j}(s_i) = \left[x_i - x_j, y_i-y_j, v_j, a_j, \mathtt{waited}_j\right],
\]
where $\mathtt{waited}$ is the accumulated time since the actor stopped at the stop sign. We also append the predicates associated with each actor. The complete feature set for vehicle $i$ is thus $\phi = (\phi_x, \phi_w, \phi_{\{-i\}})$, where the notation $\{-i\}$ should be understood as the set of all indices but $i$. All options were learned on a computer with an Nvidia GeForce GTX970M GPU. Training took approximately 12 hours.

\section{Results}
To validate the planning approach, we generated 100 random worlds in a new environment. Each world contains 0-5 other vehicles. 
These environments include more vehicles than training environments for individual options, and have the possibility of having many vehicles in the same lane.
In addition, we test in a case where there are 0-5 random vehicles, plus one vehicle stopped somewhere in the lane ahead.
The stopped car is a more challenging case, potentially requiring merging into traffic in the other lane in order to proceed. In all cases, the vehicle needs to be able to negotiate an intersection despite the presence of other vehicles.
We compare these two cases in Table~\ref{table:comparison}.



\begin{figure}[bt]
\centering
\includegraphics[width=0.99\columnwidth]{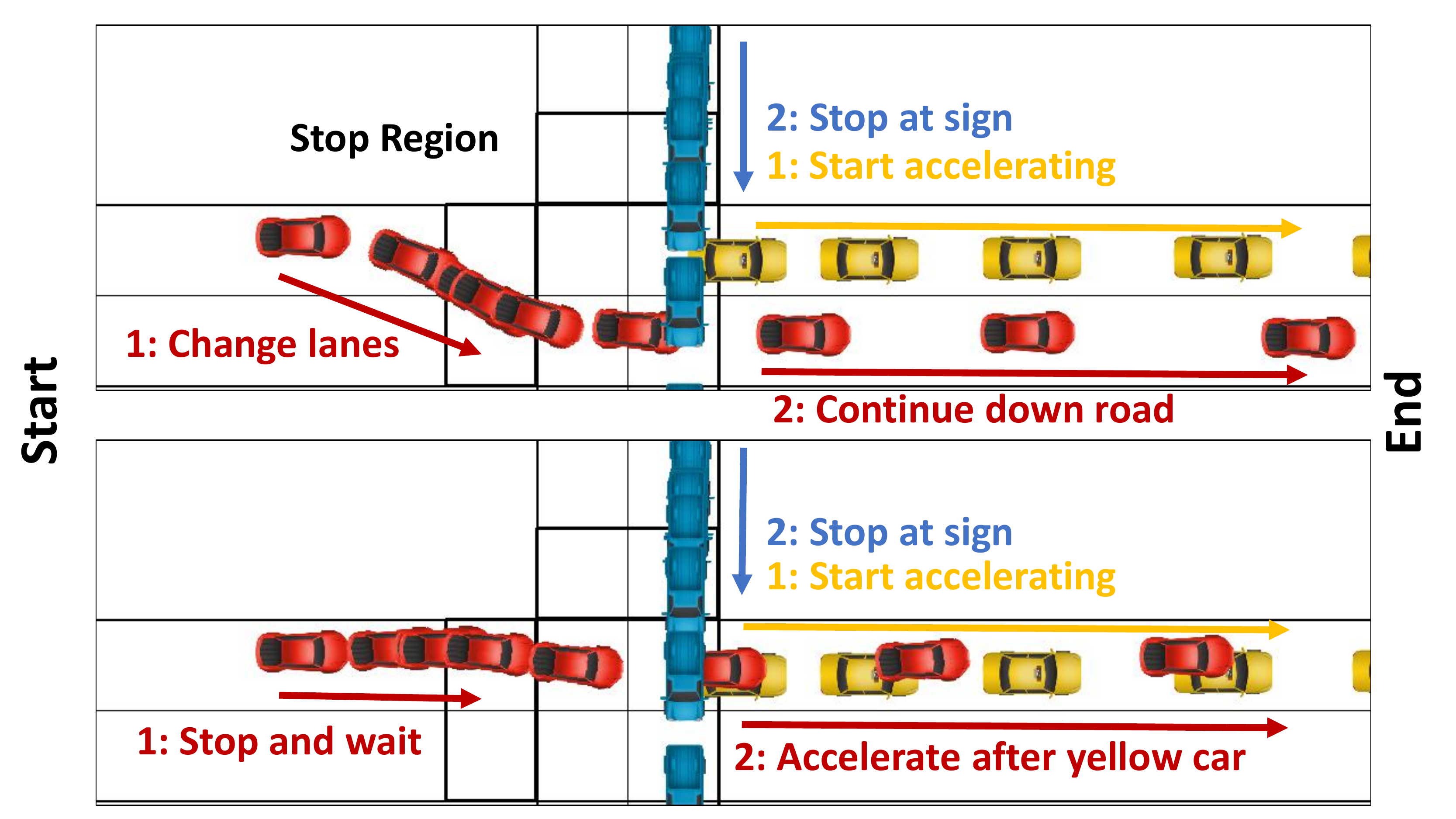}
\caption{Two different solutions to safely passing through an intersection. The red car is using the proposed algorithm. Top: manual high-level policy changes lanes, bottom: learned high-level policy waits for car in front.}
\label{fig:compare-high-level}
\end{figure}



\begin{table*}[bt]
\centering
\caption{Comparison of different algorithms on 100 randomly-generated road environments with multiple other vehicles and with a stopped car on the road ahead.}
\begin{tabular}{| c | c | c | c c c c c |}
\hline
Environment & Low-Level & High-Level & Constraint Violations & Collisions & Total Failures & Avg Reward & Std Dev Reward \\
\hline \hline
No Stopped Car & Manual Policy & None & 0 & 0 & 0 & 117.8 & 60.4 \\
& Simple & Manual & 0 & 5 & 7 & 105.5 & 93.0 \\
& Simple & Learned & 0 & 5 & 9 & 108.2 & 102.7\\
& 32 NN Policy & None & 3 & 1 & 4 & 103.2 & 108.6 \\
& 256 NN Policy & None & 3 & 4 & 6 & 109.4 & 95.7 \\
& Learned & Manual & 0 & 4 & 4 & 124.2 & 97.8 \\
& Learned & Learned & 0 & 0 & 0 & 137.8 & 74.2 \\ 
\hline
Stopped Car Ahead & Manual Policy & None & 0 & 19 & 19 & 27.2 & 142.1 \\
& Manual & Uniform & 0 & 25 & 34 & 9.3 & 162.7 \\
& Manual & Learned & 1 & 27 & 36 & 7.4 & 163.2 \\
& 32 NN Policy & None & 6 & 39 & 45 & -51.0 & 143.8 \\
& 256 NN Policy & None & 4 & 29 & 33 & -9.1 & 149.3 \\
& Learned & Manual & 1 & 9 & 11 & 83.7 & 102.2 \\ 
& Learned & Learned & 0 & 3 & 3 & 95.2 & 74.2 \\  
\hline
\end{tabular}
\label{table:comparison}
\end{table*}


\begin{figure}
\centering
\includegraphics[width=0.46\columnwidth]{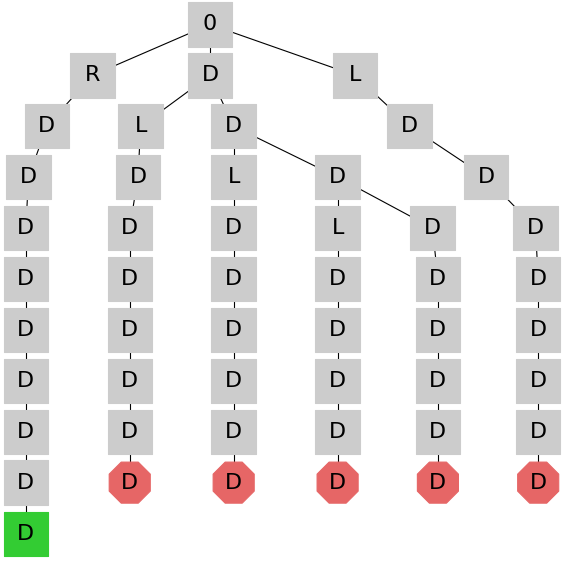}
\hskip 5mm
\includegraphics[width=0.46\columnwidth]{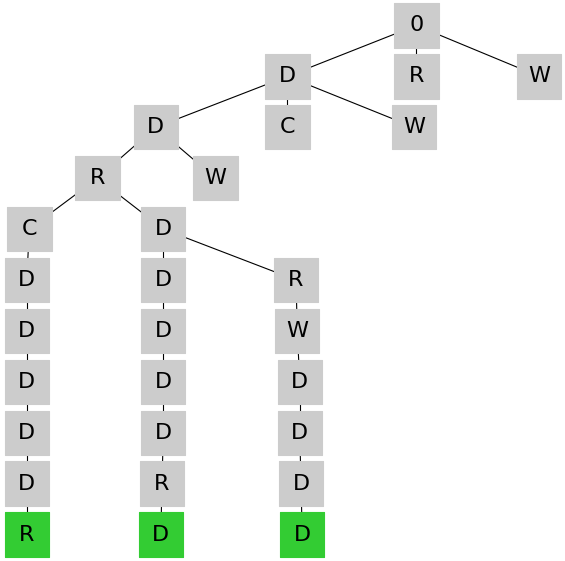}
\caption{Comparison of MCTS on a test problem with a stopped car. Letters indicate option being executed: 0 = root, D = default ``stay in lane'' policy, W = wait, C = Finish/complete level, R = lane change to the right. On the left, we see tree search with a manually defined preference; on the right, we see the tree using the high-level policy acquired through DQN. Green leaves indicate success; red leaves indicate failure.}
\label{fig:mcts}
\end{figure}

For cases with the learned or simple action sets, we performed $100$ iterations of MCTS as per Alg.~\ref{alg:complete} to a depth of $10$ seconds and select the best path to execute.
Fig.~\ref{fig:compare-high-level} and Fig.~\ref{fig:mcts} show how this functions in practice: the algorithm selects between its library of learned options, choosing to take an action that will move into a lane of moving traffic instead of into a lane where there is a stopped vehicle.

The ``manual'' policy in Table~\ref{table:comparison} executes the same aggressive policy as all the other actors on the road. It will usually come to a safe stop behind a stopped vehicle. By contrast, the ``Simple'' action set contains actions for turning to the left or right while maintaining a constant velocity, for stopping either comfortably, abruptly, or at the next goal, and for accelerating to the speed limit. It does not have any specific policy for handling LTL constraints. The manual driving policy used by other actors did better than cases that relied on MCTS over simple manual options to solve difficult problems, but not as good as cases that could fall back on learned options.

The version of the planning algorithm (with learned actions but without DQN) records a few collisions even in relatively simple problem with no stopped car. These situations occur in a few different situations, usually when the vehicle is ``boxed in'' by vehicles on multiple sides and needs to slow down.
Without the high-level policy to guide exploration, it does not know the correct sequence of options to avoid collisions.

By contrast, with the learned high-level policy, we see perfect performance on the test set for simple problems and three failures in complex problems. Note that these failures universally represent cases where the vehicle was trapped: there is a car moving at the same speed in the adjacent lane and a stopped car ahead.
Given this situation, there is no good option other than to hit brakes as hard as possible.
When our system does not start in such a challenging position, it will avoid getting into such a situation; if it predicts that such a system will arise in the next ten seconds, our planner would give us roughly 2 seconds of warning to execute an emergency stop and avoid collision.

Fig.~\ref{fig:mcts} shows a comparison of a subset of the planner calls for the version with the manually defined vs. learned high level policy. The manual policy has a preference for the default, ``stay in lane'' policy in all situations. This leads it to a number of unnecessary collisions before it finds a good solution. Fig.~\ref{fig:compare-high-level} shows the selected trajectories associated with these trees.

Our Python implementation takes roughly one second to perform a single search. As shown in Fig.~\ref{fig:system}, we expect in practice to perform one search every second to determine the set of options to execute until the next planning cycle.
A future implementation of this planner on the vehicle would operate in parallel, and would be more highly optimized. Roughly 25\% of the current speed is due to costs associated with updating the world and evaluating policies for other actors, for example, which we expect would be easily reduced with a more efficient implementation.

\section{Conclusions}\label{sec:conc}
We laid out a framework for using learned models of skills to generate task and motion plans while making minimal assumptions as to our ability to collect data and provide structure to the planning problem.
Our approach allows simple, off-the-shelf Deep Reinforcement Learning techniques to generalize to challenging new environments, and allows us to verify their behavior in these environments.

There are several avenues for improvement. First, the low-level policies learned in the self-driving car example are not perfect and sometimes can have oscillatory behavior that requires further investigation.
Second, we use termination conditions based on fixed time during tree search. Other, more complex conditions are possible under the proposed framework and should be investigated.
Additionally, choosing the set of options is still done manually, and choosing the best set is still an open problem. Finally, in our examples we use a manually-determined feature set, and it is not clear what the best set of features is.

In the future, we wish to extend this work to use stochastic control policies. These can then be combined with continuous-space MCTS improvements to perform a more inclusive search over possible trajectories. We will also apply our system to real robots.


\bibliographystyle{plainnat}
\bibliography{planners,task_description,taskmodels,drl,software,oac,ltl}

\end{document}